\documentclass[conference]{IEEEtran}
\IEEEoverridecommandlockouts
\usepackage{cite}
\usepackage{amsmath,amssymb,amsfonts}
\usepackage{algorithmic}
\usepackage{graphicx}
\usepackage{url}
\usepackage{textcomp}
\usepackage{booktabs}
\usepackage{multirow}
\usepackage{xcolor}
\def\BibTeX{{\rm B\kern-.05em{\sc i\kern-.025em b}\kern-.08em
    T\kern-.1667em\lower.7ex\hbox{E}\kern-.125emX}}
\begin{document}

\title{FTimeXer: Frequency-aware Time-series
Transformer with  Exogenous variables for Robust
Carbon Footprint Forecasting\\

\thanks{This work was supported by the Tianchi Talents - Young Doctor Program (5105250183m),  Science and Technology Program of Xinjiang Uyghur Autonomous Region (2025B04051, 2024B03028), Regional Fund of the National Natural Science Foundation of China (202512120005). Qingzhong Li, Yue Hu, Zhou Long, Qingchang Ma, Hui Ma, and Jinhai Sa are with Xinjiang Key Laboratory of Intelligent Computing and Smart Applications, School of Software, Xinjiang University, Urumqi 830046, China.}
}

\author{\IEEEauthorblockN{1\textsuperscript{st} Qingzhong~Li}
\IEEEauthorblockA{\textit{School of Software, Xinjiang University}\\
Urumqi, China\\
liqingzhong@stu.xju.edu.cn}
~\\
\and
\IEEEauthorblockN{2\textsuperscript{nd} Yue~Hu}
\IEEEauthorblockA{\textit{School of Software, Xinjiang University}\\
Urumqi, China\\
20232501422@stu.xju.edu.cn}
\and
\IEEEauthorblockN{3\textsuperscript{rd} zhou~Long}
\IEEEauthorblockA{\textit{School of Software, Xinjiang University}\\
Urumqi, China\\
longzhou@stu.xju.edu.cn}
\and
\IEEEauthorblockN{4\textsuperscript{th} Qingchang~Ma}
\IEEEauthorblockA{\textit{School of Software, Xinjiang University}\\
Urumqi, China\\
mackm@stu.xju.edu.cn}
\and
\IEEEauthorblockN{5\textsuperscript{th} Hui~Ma\textsuperscript{*}}
\IEEEauthorblockA{\textit{School of Software, Xinjiang University}\\
Urumqi, China\\
huima@xju.edu.cn\\
\textsuperscript{*}Corresponding author}
\and
\IEEEauthorblockN{6\textsuperscript{th} Jinhai~Sa}
\IEEEauthorblockA{\textit{School of Software, Xinjiang University}\\
Urumqi, China\\
super\_sjh@163.com}
}

\maketitle

\begin{abstract}
Accurate and up-to-date forecasting of the power grid's carbon footprint is crucial for effective product carbon footprint (PCF) accounting and informed decarbonization decisions. However, the carbon intensity of the grid exhibits high non-stationarity, and existing methods often struggle to effectively leverage periodic and oscillatory patterns. Furthermore, these methods tend to perform poorly when confronted with irregular exogenous inputs, such as missing data or misalignment. To tackle these challenges, we propose FTimeXer, a frequency-aware time-series Transformer designed with a robust training scheme that accommodates exogenous factors. FTimeXer features an Fast Fourier Transform (FFT)-driven frequency branch combined with gated time-frequency fusion, allowing it to capture multi-scale periodicity effectively. It also employs stochastic exogenous masking in conjunction with consistency regularization, which helps reduce spurious correlations and enhance stability. Experiments conducted on three real-world datasets show consistent improvements over strong baselines. As a result, these enhancements lead to more reliable forecasts of grid carbon factors, which are essential for effective PCF accounting and informed decision-making regarding decarbonization.
\end{abstract}



\section{Introduction}
Climate change presents systemic risks that accelerate the global transition to low-carbon energy and power systems. In response, China has announced targets to peak carbon emissions before 2030 and achieve carbon neutrality before 2060~\cite{lee2023climate}. Against this backdrop, under the "dual-carbon" goals and increasingly stringent green trade regulations, electricity-related emissions from enterprises and industrial parks (Scope 2) have become a crucial component of carbon accounting, low-carbon operations, and supply chain compliance disclosure. Consequently, the temporal dynamics of grid carbon footprints, including carbon emissions and carbon intensity, directly impact the accuracy and timeliness of product carbon footprint accounting as well as the assessment of mitigation measures~\cite{European2023regulation,protocol2015ghg,maji2022dacf}.


However, achieving high-accuracy forecasting for power grid carbon emissions still faces two key challenges. Although prior studies have explored statistical methods, traditional machine learning approaches, and deep learning techniques for short-term forecasting and assessment of power system emissions or carbon intensity~\cite{ma2022tsdcnn,xi2023cvd,wang2024polysilicon_scheduling}, most existing methods remain primarily time-domain based. Therefore, these methods fail to adequately capture the periodicity, oscillatory behavior, and multi-scale spectral structures in grid carbon emission series, which are often induced by intra-day load cycles, weekly cycles, and unit-commitment switching, thereby degrading the model's generalization performance.
~\cite{ye2024fan}. 


In addition, practical energy consumption forecasting is also significantly influenced by exogenous conditions~\cite{exogfusion,olivares2023nbeatsx}. Therefore, even a frequency-aware temporal model can become unreliable when the external signals it relies on are imperfect or temporally misaligned, leading to degraded generalization performance~\cite{wei2023coformer}. Existing energy and emission forecasting methods commonly treat exogenous factors, such as electricity prices, meteorological conditions, operational states, and scheduling signals, as auxiliary inputs. They enhance performance through techniques like feature concatenation, attention-based fusion, or straightforward multivariate modeling~\cite{tft,nbeatsx}. However, in real industrial settings, such exogenous data often suffer from heterogeneous sampling rates, missing values, and temporal misalignment. This results in them being partially observed and uncertain~\cite{tipirneni2022sst,tashiro2021csdi,du2023saits}. As a result, directly inputting naively aligned exogenous sequences into forecasting models can cause noise and misalignment to be mistaken for informative signals, resulting in spurious correlations~\cite{huang2025exollm}. When models heavily rely on exogenous inputs, these irregularities can significantly undermine forecasting stability and generalization across different scenarios.


To address these challenges, we propose FTimeXer, a frequency-enhanced time-series forecasting model designed for energy consumption modeling at key stages of silicon-based manufacturing. Specifically, we incorporate an FFT-driven frequency-domain branch into a Transformer-based temporal modeling backbone and fuse it with the time-domain branch using a gated mechanism, thereby enhancing the representation of periodic and oscillatory patterns and resulting in improved forecasting accuracy under complex fluctuating conditions. In addition, to enhance robustness and generalization to incomplete exogenous inputs without modifying the backbone architecture, we introduce a robust training strategy for irregular exogenous variables. This strategy involves applying stochastic exogenous masking to simulate missing or uncertain observations and imposing a consistency regularization term to encourage stable predictions under exogenous perturbations.

The main contributions of this work can be summarized as follows:
\begin{itemize}
\item To the best of our knowledge, we present an energy forecasting framework that incorporates an FFT-driven frequency branch with gated time-frequency fusion, which explicitly enhances the representation of multi-scale periodic fluctuations.
\item We develop a robustness scheme for irregular exogenous variables that combines stochastic exogenous masking and consistency regularization to suppress spurious correlations and enhance generalization under missing and misaligned inputs.
\item Finally, experiments conducted on three real-world datasets demonstrate that our method achieves superior performance and offers reliable predictive support for carbon footprint accounting and emission-aware optimization.
\end{itemize}

The remainder of this paper is organized as follows: Section II reviews related work; Section III presents the proposed methodology; Section IV reports experimental results; and Section V concludes the paper.

\section{RELATED WORK}
\subsection{Carbon Footprint Forecasting}
Carbon footprint forecasting is a domain-specific time-series task within power systems. Early studies primarily relied on statistical models, such as ARIMA, and conventional machine learning methods, including support vector regression (SVR), random forests, and gradient boosting. These approaches utilized hand-crafted temporal features and exogenous variables, including weather conditions, electricity demand, and generation mix. More recently, deep learning approaches, including recurrent neural networks (RNN), long short-term memory networks (LSTM), gated recurrent units (GRU), and temporal convolutional networks (TCN), have been introduced to enhance non-linear modeling capacity and robustness in the context of non-stationary emissions dynamics.

In recent years, Transformer architectures and their variants have demonstrated remarkable performance in time-series forecasting. Representative works enhance scalability for long sequences through sparse and efficient attention mechanisms, such as Informer~\cite{informer} and ETSformer~\cite{etsformer}. They explicitly model trend and seasonal components by employing series decomposition methods, exemplified by Autoformer~\cite{autoformer}. Additionally, these works further improve representation learning and training efficiency using patch-based modeling or structural reparameterization techniques, as demonstrated by PatchTST~\cite{patchtst} and iTransformer~\cite{itransformer}. These methods have made significant advancements in general long-horizon forecasting tasks.

However, for industrial energy and grid-carbon signals, purely time-domain modeling is often inadequate for capturing multi-scale periodic fluctuations and oscillatory structures. Moreover, generalization tends to deteriorate further under noise perturbations and error accumulation in long-horizon forecasting.

\subsection{Fourier-transform-driven frequency-domain modeling}
Frequency-domain methods offer a complementary perspective by explicitly characterizing periodicity and oscillations. Classical studies typically employ Fourier or wavelet transforms, as well as decomposition methods, for denoising and extracting periodic components. More recently, many forecasting models have integrated frequency-domain representations or spectral operators into deep learning frameworks to enhance the learning of periodic patterns. A comprehensive survey can be found in the following reference~\cite{freqsurvey}. For example, Zhou \textit{et al.}~\cite{zhou2022fedformer} proposed FEDformer, a model that incorporates FFT-based spectral sampling and frequency-domain attention into Transformers to facilitate efficient long-horizon forecasting. Cao \textit{et al.}~\cite{stemgnn} proposed StemGNN for multivariate graph time series, leveraging a graph Fourier basis along with time-frequency modeling to capture cross-variable dependencies and temporal periodicity. Li \textit{et al.}~\cite{fno} proposed the Fourier Neural Operator (FNO), a model that learns operators in the frequency domain to enhance the modeling of long-range dependencies.

However, in many existing approaches, frequency information is treated as an auxiliary feature. Moreover, the interaction between temporal and spectral representations lacks an explicit, controllable, and adaptive fusion mechanism, which can lead to degraded forecasting accuracy under complex fluctuations and disturbances.

\subsection{Exogenous modeling and robust learning}
To enhance forecasting performance, many studies treat exogenous factors—such as electricity prices, meteorological conditions, operational states, and scheduling signals—as auxiliary inputs. These factors are typically integrated using methods such as feature concatenation, attention-based fusion, or conditional injection~\cite{exogfusion}. Meanwhile, robustness-oriented techniques, including masking-based augmentation, perturbation-robust training, and consistency regularization, have also been explored in time-series and representation learning~\cite{masking,consistency}.

However, despite these efforts, most exogenous-aware forecasting still assumes that external signals are well-aligned and reliable. Under real-world conditions of missing data, noise, and temporal misalignment, models can easily learn spurious correlations and may experience degraded stability. As a result, a backbone-compatible robust training paradigm specifically designed for irregular exogenous inputs remains underexplored.



\begin{figure*}[t]
  \centering
   \includegraphics[width=0.85\linewidth]{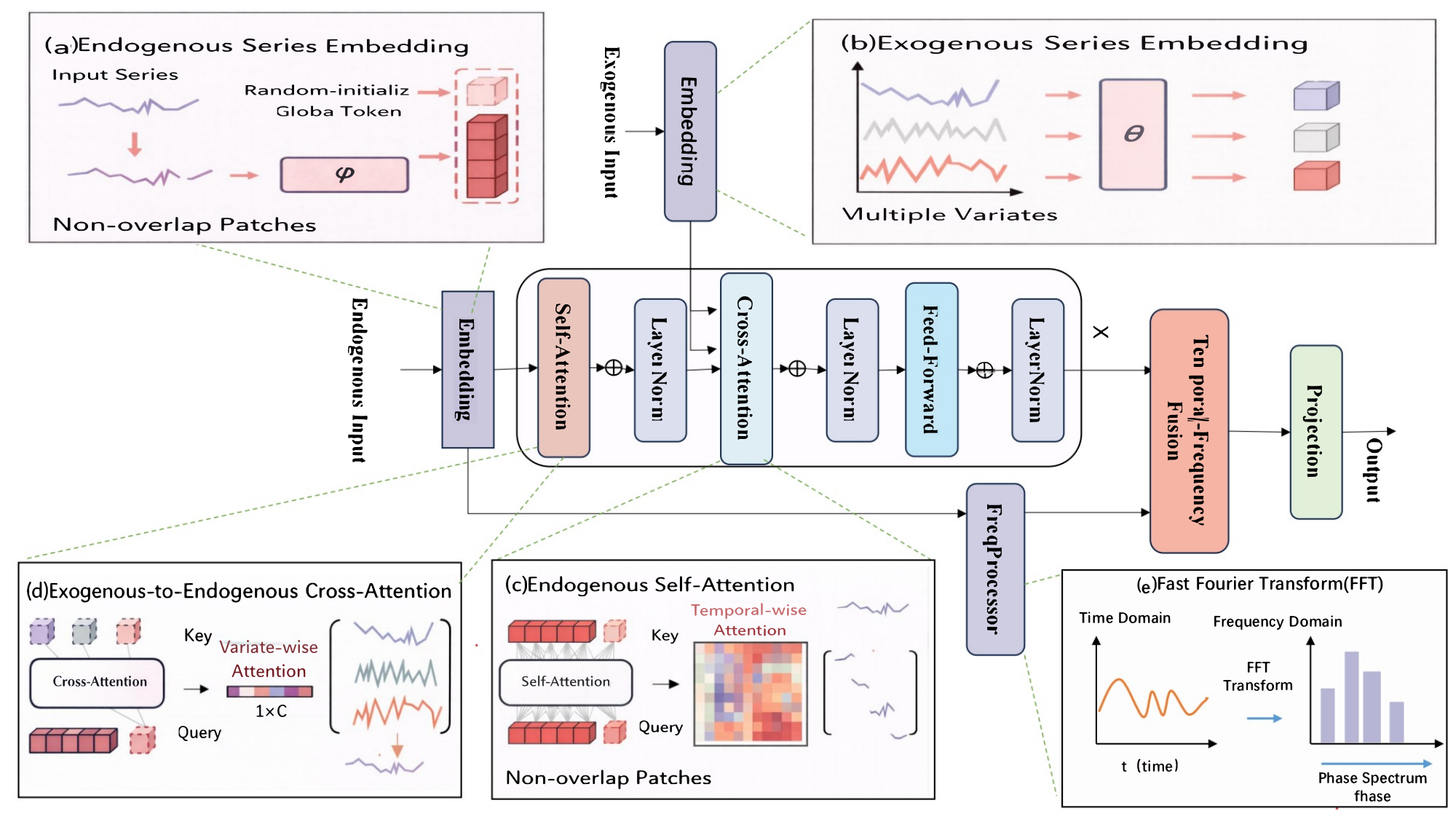}
   \caption{Structure of FTimeXer model.
   }
   \label{fig:framework}
\end{figure*}
\section{METHOD}
\subsubsection{Overall Framework}
Based on the standard Transformer architecture for time-series modeling, we design a time-frequency fusion forecasting framework. namely, Frequency-aware Time-series Transformer with Exogenous variables for Robust Carbon Footprint Forecasting (FTimeXer). The proposed framework consists of three core components: a time-domain modeling branch, a frequency-domain modeling branch, and a fusion mechanism. Specifically, the time-domain branch captures long-range temporal dependencies, while the frequency-domain branch extracts latent periodic and oscillatory patterns using the Fast Fourier Transform (FFT). These two branches are then integrated through a gated fusion module, enabling collaborative modeling of time-frequency information. For clarity, the overall framework is illustrated in Fig.~\ref{fig:framework}.

\subsection{Problem Formulation}

\subsubsection{Endogenous Sequence Embedding}

Most existing Transformer-based forecasting models embed each time point or segment of a time series as a temporal token and apply self-attention mechanisms to learn temporal dependencies. To accurately capture the temporal variations within endogenous variables, our model employs a patch-based representation. Specifically, the endogenous sequence is divided into non-overlapping blocks, with each block being projected as a temporal token.

The input of endogenous variables is represented as a multivariate time series $\mathbf{X}^{(e)} = [x_1, x_2, \dots, x_T] \in \mathbb{R}^{T \times d_e}$, where $T$ denotes the sequence length (number of time steps) and $d_e$ represents the feature dimension of the endogenous variables. This sequence is sliced into several non-overlapping patches, with each patch containing data from consecutive time steps. Through an embedding function $\text{PatchEmbed}(\cdot)$—typically a linear projection or a 1-D convolution—each patch is mapped to a high-dimensional representation, i.e., a patch token.

Considering the distinct roles of endogenous and exogenous variables in forecasting, the model embeds them at different granularities. Directly combining endogenous tokens and exogenous tokens of different granularities would lead to information misalignment. To address this issue, we introduce a learnable global token for each endogenous variable, denoted as $\emptyset \in \mathbb{R}^{1 \times d}$, where $d$ is the hidden dimension of the model. This global token serves as a macroscopic representation used to interact with exogenous variables, thereby bridging the causal information of the exogenous series into the endogenous time patches. Finally, the embedding representation of the endogenous series can be formulated as:

\begin{equation}
Z_0^{(e)}=\left[\emptyset ; \cdot \operatorname{PatchEmbed}\left(X^{(e)}\right)\right]
\end{equation}


where $\emptyset$ denotes a global token that enables cross-modal interaction with exogenous variables in the subsequent attention mechanism, and $PatchEmbed(\cdot)$ is the patch embedding function (implemented via a linear mapping or a 1D convolution) that maps each non-overlapping time block to a token representation.

\subsubsection{Exogenous Sequence Embedding}

The primary role of exogenous variables is to provide auxiliary and interpretable background information for predicting endogenous variables, such as weather conditions, energy prices, equipment status, and macroscopic policy signals. However, unlike endogenous variables, exogenous data often exhibit significant irregularities, including timestamp misalignment, varying sampling frequencies, missing values, and inconsistent look-back lengths. Consequently, adopting the same patch embedding strategy for exogenous variables as that used for endogenous variables would not only introduce substantial noise irrelevant to the prediction target but also significantly increase computational complexity due to overly fine-grained partitioning. This, in turn, could obscure high-level semantic associations between variables. Therefore, we advocate for a more adaptive representation method for exogenous variables.

To achieve this, we adopt a variable-level embedding strategy. Specifically, each exogenous variable sequence is mapped holistically to a single, globally representative variable token. Formally, given the exogenous variable input tensor $X^{(x)} \in \mathrm{R}^{T \times d_x}$, where $T$ represents the total number of time steps aligned with the endogenous sequence and $d_x$ represents the original feature dimension of the exogenous variables, we project the entire sequence into a feature space whose dimensionality is consistent with the model's hidden layer through a shared linear transformation layer:

\begin{equation}Z_0^{(x)}=X^{(x)}\theta+b,\theta\in R^{d_x\times d},b\in{R}^d
\end{equation}

In this formula, $\mathbf{\theta}$ is a learnable weight matrix responsible for mapping the exogenous features at each time step from dimension $d_x$ to the model's hidden dimension $d$, while $\mathbf{b}$ is the corresponding bias vector. This linear projection provides a preliminary feature transformation for each exogenous variable at the time-step level. However, to obtain a compact sequence-level representation for subsequent cross-variable attention interactions, we typically perform an aggregation operation on $\mathbf{Z}_0^{(x)}$ along the temporal dimension $T$. In practice, common aggregation methods include global average pooling and a lightweight attention pooling layer.Finally, we obtain a variable token for each exogenous variable, denoted as $\mathbf{v}_j \in \mathrm{R}^d$, where $j$ indexes the different exogenous variables. The tokens of all exogenous variables collectively form the exogenous embedding set, which is used to exchange information with the global token of the endogenous sequence. This design enables the model to flexibly and efficiently fuse contextual information from irregular exogenous data, thereby enhancing its understanding and predictive capability regarding the dynamic changes in endogenous variables.

\subsubsection{Temporal Self-Attention Mechanism}

For accurate time series forecasting, it is crucial to discover the inherent temporal dependencies within endogenous variables and their interactions with variable-level representations derived from exogenous variables. Beyond the self-attention on endogenous temporal tokens (block-to-block), the learnable global token serves as a bridge between endogenous and exogenous variables. Specifically, the global token plays an asymmetric role in the cross-attention mechanism: (1) Block-to-Global: the global token attends to temporal tokens to aggregate block-level information from the entire sequence; (2) Global-to-Block: each temporal token attends to the global token to capture variable-level correlations. As a result, this design provides a comprehensive perspective on the internal temporal dependencies of endogenous variables and facilitates better interaction with arbitrary irregular exogenous variables. Accordingly, the modeling of temporal dependencies for the endogenous embedding $\mathbf{Z}_l^{(e)}$ is represented as follows:

\begin{equation}
    Q = \mathbf{Z}_l^{(e)} \mathbf{W}_Q, \quad K = \mathbf{Z}_l^{(e)} \mathbf{W}_K, \quad V = \mathbf{Z}_l^{(e)} \mathbf{W}_V
\end{equation}

To enable each temporal token of the endogenous sequence to focus on its own temporal context while selectively absorbing macroscopic semantic information from exogenous variables via the global token as a mediator, we update the endogenous representation through a cross-attention mechanism. As a result, the contextual features of the endogenous sequence are enhanced in the temporal dimension, providing a richer representation for subsequent fine-grained predictions. The updated result is as follows:

\begin{equation}
    \tilde{\mathbf{Z}}_l^{(e)} = \mathbf{Z}_l^{(e)} + \text{CrossAttn}(\mathbf{Z}_l^{(e)}, \mathbf{Z}_l^{(x)})
\end{equation}


\subsubsection{Exogenous-Endogenous Cross-Attention Mechanism}

Cross-attention is widely used in multimodal learning to capture adaptive token-level dependencies between different modalities. In our model, the cross-attention layer uses endogenous variables as Queries and exogenous variables as Keys and Values to couple these two types of variables. Since exogenous variables are embedded as variable-level tokens, we utilize the learned global token of the endogenous variables to aggregate information from the exogenous variables. Formally, we take the exogenous embedding $\mathbf{Z}_l^{(x)}$ as Key/Value and the endogenous embedding $\tilde{\mathbf{Z}}_l^{(e)}$ as Query:

\begin{equation}
    Q = \tilde{\mathbf{Z}}_l^{(e)} \mathbf{W}_Q, \quad K = \mathbf{Z}_l^{(x)} \mathbf{W}_K, \quad V = \mathbf{Z}_l^{(x)} \mathbf{W}_V
\end{equation}

To further integrate the auxiliary information provided by exogenous variables with the temporally contextualized representation, we aim to enable the model to simultaneously capture the synergistic effects of endogenous sequence patterns and exogenous conditions, thereby improving the accuracy and interpretability of predictions. Accordingly, we update the cross-variable correlations. The representation updated with cross-variable correlations is obtained as follows:

\begin{equation}
    \hat{\mathbf{Z}}_l^{(e)} = \tilde{\mathbf{Z}}_l^{(e)} + \text{CrossAttn}(\mathbf{Z}_l^{(e)}, \mathbf{Z}_l^{(x)})
\end{equation}



\subsubsection{Frequency Domain Processing Module}

We transform the endogenous sequence from the time domain to the frequency domain to extract spectral features effectively. For the endogenous time-block token sequence at the $l$-th layer, we first pass it through a linear projection to obtain a pseudo-time-domain representation that is more suitable for frequency-domain transformation. Next, a one-dimensional Fast Fourier Transform (FFT) is performed independently on each feature channel to transform it from the time domain to the frequency domain:

\begin{equation}
    F(\mathbf{X}^{(e)}) = \text{FFT}(\mathbf{X}^{(e)}) = \{A(f), \varphi(f)\}
\end{equation}

 Here, $A(f)$ represents the amplitude spectrum, which characterizes the energy distribution of the signal across different frequency components. Correspondingly, $\varphi(f)$ represents the phase spectrum, which describes the temporal structure of the signal across its frequency components.

A frequency-domain processing layer, such as a multilayer perceptron (MLP) or a convolutional layer, is used to extract both high- and low-frequency components. Such processing enables the model to learn and emphasize important frequency components while simultaneously suppressing noise:

\begin{equation}
    F_{freq} = \sigma(\mathbf{W}_f \cdot A(f) + \mathbf{b}_f)
\end{equation}

Here, $\mathbf{W}_f$ and $\mathbf{b}_f$ are learnable parameters, while $\sigma$ denotes the activation function. This frequency-domain processing step can be flexibly designed to extract high-frequency details or to smooth low-frequency trends. Next, the processed frequency-domain feature $F{freq}^{(l)}$ is combined with the original phase spectrum $\varphi^{(l)}(f)$ and then reconstructed back to the time-domain representation using the Inverse Fast Fourier Transform (IFFT):

\begin{equation}
    \tilde{\mathbf{X}}^{(e)} = \text{Re}[\text{IFFT}(F_{freq} e^{j\varphi(f)})]
\end{equation}

Finally, $\tilde{\mathbf{X}}^{(e)}$ is re-mapped back to the token dimension through a linear projection to obtain the output of the frequency domain path.

\subsubsection{Time-Frequency Fusion Module}

This module aims to effectively fuse the time-domain output from the model's main path with the output from the frequency-domain path. Specifically, we combine the endogenous time-block token output from the $l$-th layer with the frequency-domain output. Accordingly, the fusion of the time-domain output $\hat{\mathbf{Z}}_l^{(e)}$ and the frequency-domain output $\tilde{\mathbf{X}}^{(e)}$ can be represented as follows:

\begin{equation}
    \mathbf{Z}^{(fusion)} = \text{MLP}([\hat{\mathbf{Z}}_l^{(e)} \parallel \tilde{\mathbf{X}}^{(e)}])
\end{equation}

Here, $[\cdot \parallel \cdot]$ denotes concatenation along the feature dimension, and the multilayer perceptron (MLP) projects the fused features back to the original dimension. Consequently, this concatenation explicitly preserves information from both the time and frequency domains and enables rich interactions between them.

\subsubsection{Feed-Forward Network and Layer Normalization}

The fused token sequence undergoes nonlinear transformation and stabilization through the Feed-Forward Network (FFN) and Layer Normalization, following the standard Transformer architecture. In this structure, each layer is computed as follows:

\begin{equation}
    \mathbf{Z}^{(ff)} = \text{FFN}[\text{LayerNorm}(\mathbf{Z}^{(fusion)})]
\end{equation}

In this context, the Feed-Forward Network (FFN) typically consists of two linear layers and an activation function. The output from this layer serves as the input endogenous time block token for the next layer, thereby integrating the frequency-domain information flow into the deeper layers of the model.
\begin{table*}[t]
\centering
\caption{Performance comparison on three datasets.}
\label{tab:three_datasets_metrics}
\small
\setlength{\tabcolsep}{2pt}
\begin{tabular*}{\textwidth}{@{\extracolsep{\fill}}lcccc cccc cccc}
\toprule
\multirow{2}{*}{Model} &
\multicolumn{4}{c}{Magnolia} &
\multicolumn{4}{c}{California\_CT2} &
\multicolumn{4}{c}{NowYork\_Greenidge} \\
\cmidrule(lr){2-5}\cmidrule(lr){6-9}\cmidrule(lr){10-13}
& $R^2\uparrow$ & MSE$\downarrow$ & RMSE$\downarrow$ & MAE$\downarrow$
& $R^2\uparrow$ & MSE$\downarrow$ & RMSE$\downarrow$ & MAE$\downarrow$
& $R^2\uparrow$ & MSE$\downarrow$ & RMSE$\downarrow$ & MAE$\downarrow$ \\
\midrule
GRU
& 0.819 & 47.552 & 6.896 & 4.681
& 0.744 & 671.380 & 25.911 & 17.911
& 0.678 & 22.816 & 4.777 & 3.320 \\
LSTM
& 0.830 & 44.796 & 6.693 & 4.627
& 0.809 & 501.571 & 22.396 & 13.993
& 0.667 & 23.630 & 4.861 & 2.913 \\
Transformer
& 0.837 & 42.880 & 6.548 & 4.423
& 0.787 & 557.731 & 23.616 & 15.068
& 0.707 & 20.817 & 4.563 & 2.575 \\
Informer~\cite{informer}
& 0.858 & 37.398 & 6.115 & 5.076
& 0.781 & 574.756 & 23.974 & 19.119
& 0.724 & 19.580 & 4.425 & 2.442 \\
TimeXer~\cite{exogfusion}
& 0.873 & 33.377 & 5.777 & 3.512
& 0.822 & 468.237 & 21.639 & 12.215
& 0.726 & 19.430 & 4.408 & 2.338 \\
FTimeXer
& \textbf{0.890} & \textbf{30.690} & \textbf{5.514} & \textbf{3.386}
& \textbf{0.843} & \textbf{420.750} & \textbf{20.414} & \textbf{11.781}
& \textbf{0.762} & \textbf{17.028} & \textbf{4.109} & \textbf{2.237} \\
\bottomrule
\end{tabular*}
\end{table*}
\subsubsection{Output Projection Layer}
\begin{table*}[t]
\centering
\caption{Ablation study on Magnolia, California\_CT2, and NewYork\_Greenidge datasets.}
\label{tab:mask_consistency_example}
\small
\setlength{\tabcolsep}{1.8pt}
\renewcommand{\arraystretch}{1.10}
\begin{tabular*}{\textwidth}{@{\extracolsep{\fill}}lcccc cccc cccc}
\toprule
\multirow{2}{*}{Setting} &
\multicolumn{4}{c}{Magnolia} &
\multicolumn{4}{c}{California\_CT2} &
\multicolumn{4}{c}{NowYork\_Greenidge} \\
\cmidrule(lr){2-5}\cmidrule(lr){6-9}\cmidrule(lr){10-13}
& $R^2\uparrow$ & MSE$\downarrow$ & RMSE$\downarrow$ & MAE$\downarrow$
& $R^2\uparrow$ & MSE$\downarrow$ & RMSE$\downarrow$ & MAE$\downarrow$
& $R^2\uparrow$ & MSE$\downarrow$ & RMSE$\downarrow$ & MAE$\downarrow$ \\
\midrule
Baseline 
& 0.879 & 33.759  & 5.782  & 3.594
& 0.816 & 498.366 & 22.216 & 12.444
& 0.747 & 18.265  & 4.247  & 2.356 \\
Masking 10\%
& 0.881 & 33.165  & 5.732  & 3.544
& 0.823 & 470.250 & 21.572 & 12.325
& 0.751 & 17.919  & 4.207  & 2.317 \\
Masking 20\%
& 0.883 & 32.472  & 5.673  & 3.505
& 0.828 & 455.400 & 21.235 & 12.177
& 0.754 & 17.672  & 4.188  & 2.297 \\
Masking 30\%
& 0.887 & 31.581  & 5.594  & 3.465
& 0.836 & 435.600 & 20.770 & 11.930
& 0.759 & 17.424  & 4.158  & 2.277 \\
Masking 40\%
& 0.884 & 32.274  & 5.653  & 3.495
& 0.831 & 450.450 & 21.117 & 12.078
& 0.756 & 17.573  & 4.168  & 2.287 \\
Masking 50\%
& 0.880 & 33.660  & 5.772  & 3.564
& 0.820 & 480.150 & 21.800 & 12.375
& 0.752 & 17.870  & 4.207  & 2.327 \\

\textbf{FTimeXer}
& \textbf{0.890} & \textbf{30.690}  & \textbf{5.514}  & \textbf{3.386}
& \textbf{0.843} & \textbf{420.750} & \textbf{20.414} & \textbf{11.781}
& \textbf{0.762} & \textbf{17.028}  & \textbf{4.109}  & \textbf{2.237} \\
\bottomrule
\end{tabular*}
\end{table*}
After processing through $L$ layers of enhanced modules, we obtain the final endogenous token representation, which includes both temporal block tokens and the global token. Finally, the prediction is generated through a linear projection as follows:

\begin{equation}
    \hat{\mathbf{Y}} = \mathbf{Z}_L \mathbf{W}_o + \mathbf{b}_o
\end{equation}
where $L$ denotes the number of layers, and $\mathbf{b}_o$ is a learnable parameter.


\subsubsection{Robust Training with Irregular Exogenous Inputs}
\label{sec:robust_training}

To enhance robustness against missing and misaligned exogenous variables in real industrial settings, we propose a backbone-agnostic robust training strategy that does not alter the model architecture.
Specifically, during training, we stochastically perturb the exogenous input and enforce prediction consistency to prevent the model from overfitting to unreliable exogenous cues.

\paragraph{Stochastic exogenous masking.}
Given the exogenous input sequence $X^{(x)}$, we randomly mask a subset of exogenous entries with a probability of $p$:
\begin{equation}
\tilde{X}^{(x)} = M \odot X^{(x)}
\end{equation}
where $M\in\{0,1\}^{T\times d_x}$ is a Bernoulli mask with $\Pr(M_{t,k}=0)=p$, and $\odot$denotes element-wise multiplication. 
This augmentation simulates the scenario of missing or partially observed exogenous signals.

\paragraph{Consistency regularization}
We further encourage the model to generate stable predictions in response to exogenous perturbations.
Let $\hat{\mathbf{Y}} = f(X^{(e)}, X^{(x)})$ represent the prediction using the original exogenous input, and let
$\hat{\mathbf{Y}}^{\prime} = f(X^{(e)}, \tilde{X}^{(x)})$ denote the prediction using the masked exogenous input. 
We define a consistency loss as follows:
\begin{equation}
\mathcal{L}_{cons} = \left\| \hat{\mathbf{Y}} - \hat{\mathbf{Y}}^{\prime} \right\|_2^2 .
\end{equation}

\paragraph{Training objective}
We first define the primary forecasting loss w.r.t. the ground truth:
\begin{equation}
\mathcal{L}_{pred}=\frac{1}{N}\sum_{i=1}^{N}\ell\!\left(\mathbf{Y}_i,\hat{\mathbf{Y}}_i\right),
\end{equation}
where $\ell(\cdot)$ is instantiated as MAE or MSE. The overall training objective is then formulated as
\begin{equation}
\mathcal{L} = \mathcal{L}_{pred} + \lambda \mathcal{L}_{cons},
\end{equation}
where $\lambda$ controls the strength of consistency regularization.

\section{Experimental Results}
\subsection{Experimental Details}
We evaluate our method using three real-world emissions time-series datasets: Magnolia, California\_CT2, and NewYork\_Greenidge. All datasets are sourced from the Clean Air Markets Division (CAMD) of the U.S. Environmental Protection Agency (EPA)\footnote{\url{https://campd.epa.gov/}}
, which publicly releases emissions-related records and facility attributes for various clean air market programs, such as the Acid Rain Program (ARP) and the Cross-State Air Pollution Rule (CSAPR).
For each site, we extract approximately three years of historical measurements. Following a chronological split, the earliest 80\% of the samples are used for training, while the remaining 20\% are reserved for testing. We consider a one-step-ahead forecasting setting in which the model uses observations from the previous 12 hours as input to predict emissions for the next hour.
\subsection{Overall Comparison}
We compare FTimeXer with representative baselines in Table~\ref{tab:three_datasets_metrics}. GRU and LSTM generally yield higher errors due to their limited capability in modeling long-range dependencies and multi-scale patterns. Transformers and Informers enhance global dependency modeling; however, they primarily operate in the time domain and do not explicitly exploit spectral periodicity. This limitation results in performance variations across different datasets. TimeXer benefits from exogenous inputs; however, it is sensitive to missing data and temporal misalignment, which may introduce spurious correlations. In contrast, FTimeXer combines an FFT-based frequency branch with gated time-frequency fusion and robust training techniques, including random exogenous masking and consistency regularization, achieving better results across all three datasets.

\subsection{Ablation Study}
Table~\ref{tab:mask_consistency_example} presents the ablation results of the proposed robust learning strategy on three real-world datasets. Specifically, this analysis covers various exogenous masking ratios and the inclusion of additional consistency regularization. Overall, introducing exogenous masking consistently improves performance compared to the baseline across all three datasets. This suggests that randomly masking external signals during training can effectively mitigate the negative impact of imperfect exogenous inputs. These results further confirm that moderate exogenous masking enhances robustness to missing data and temporal misalignment in exogenous variables. Additionally, the consistency constraint helps suppress spurious correlations, resulting in improved cross-dataset generalization.

\subsection{Visualization Analysis}
\begin{figure}[t]
  \centering
  \includegraphics[width=0.85\columnwidth]{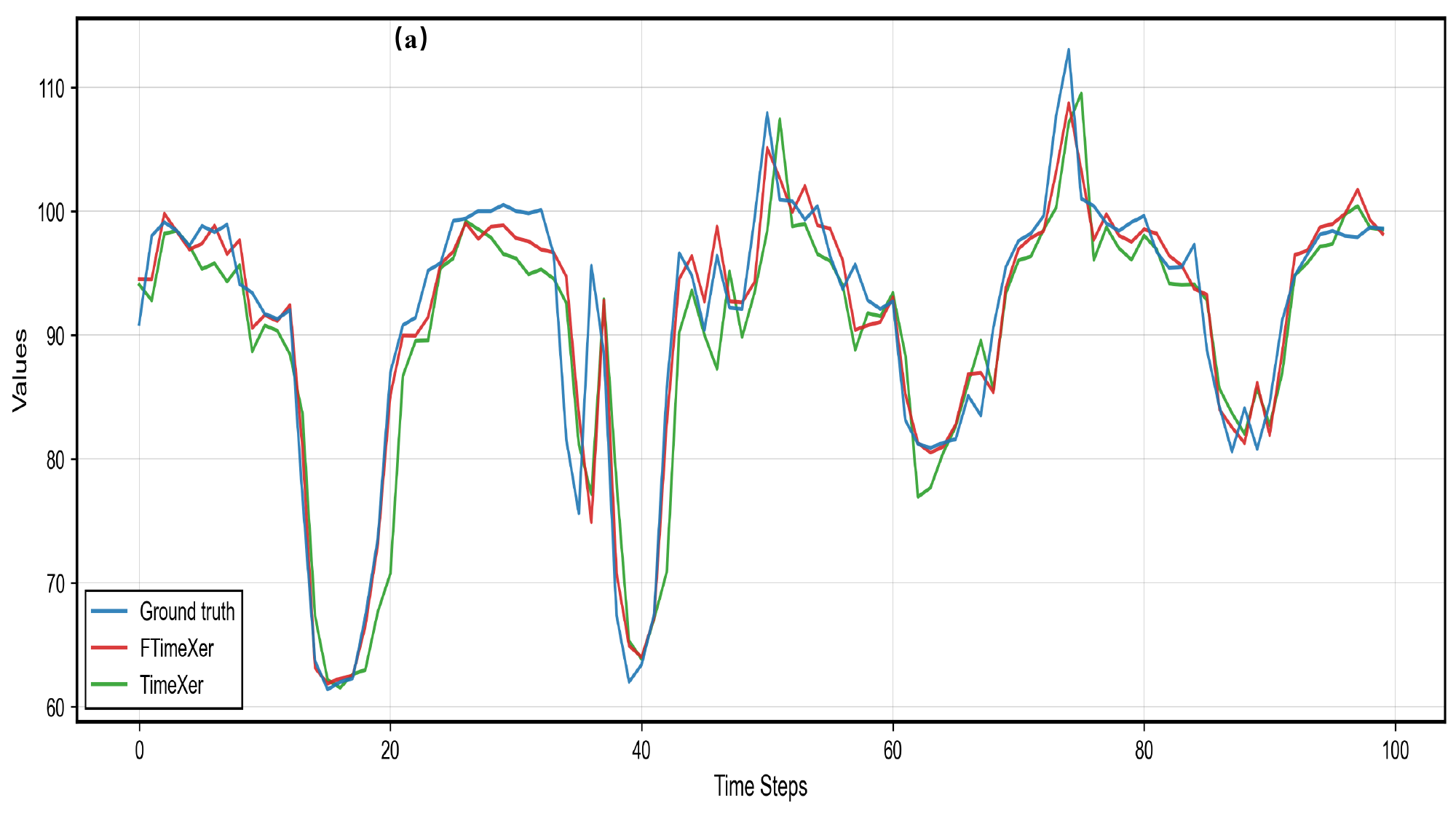}
  \includegraphics[width=0.85\columnwidth]{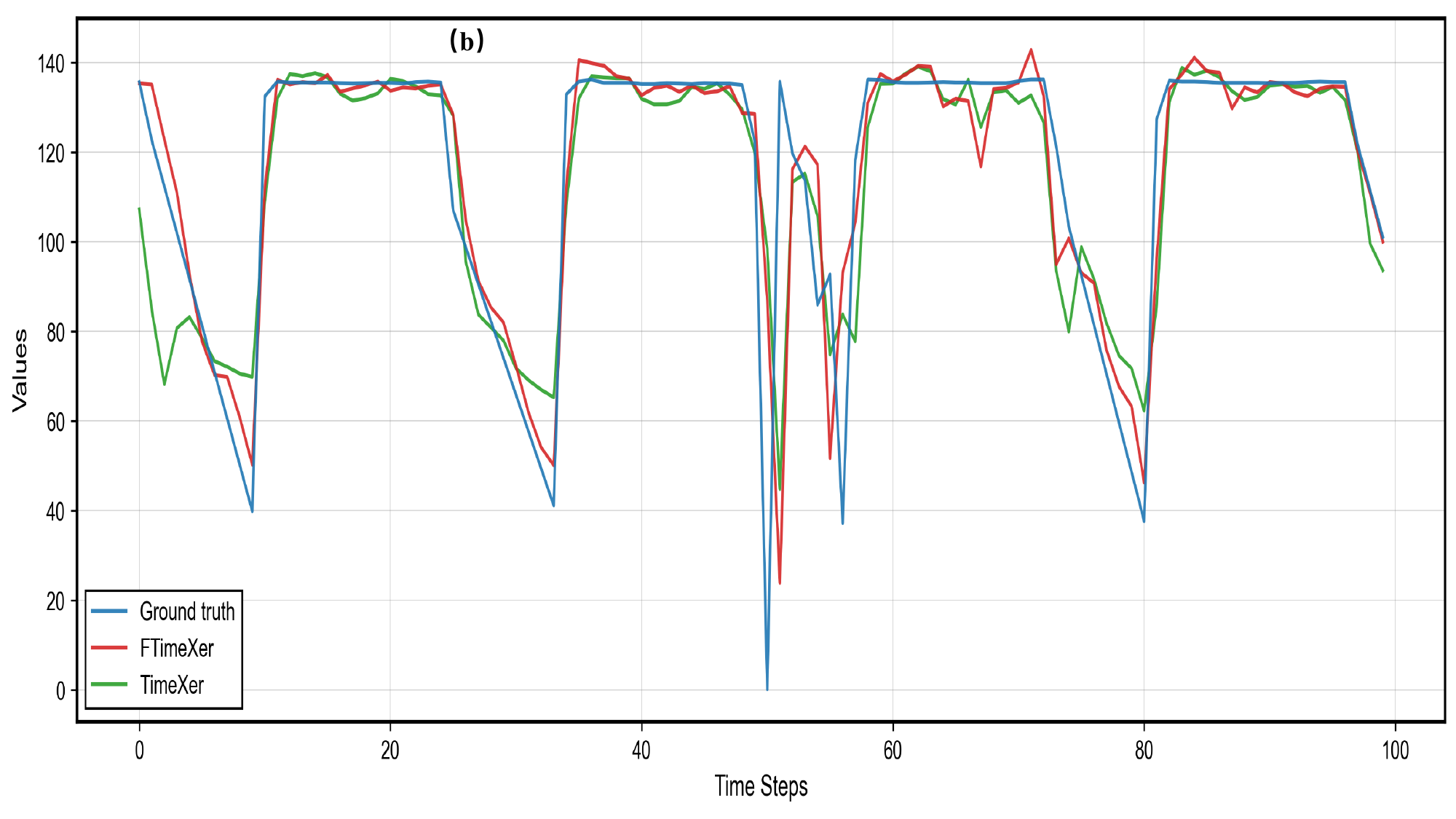}
   \includegraphics[width=0.85\columnwidth]{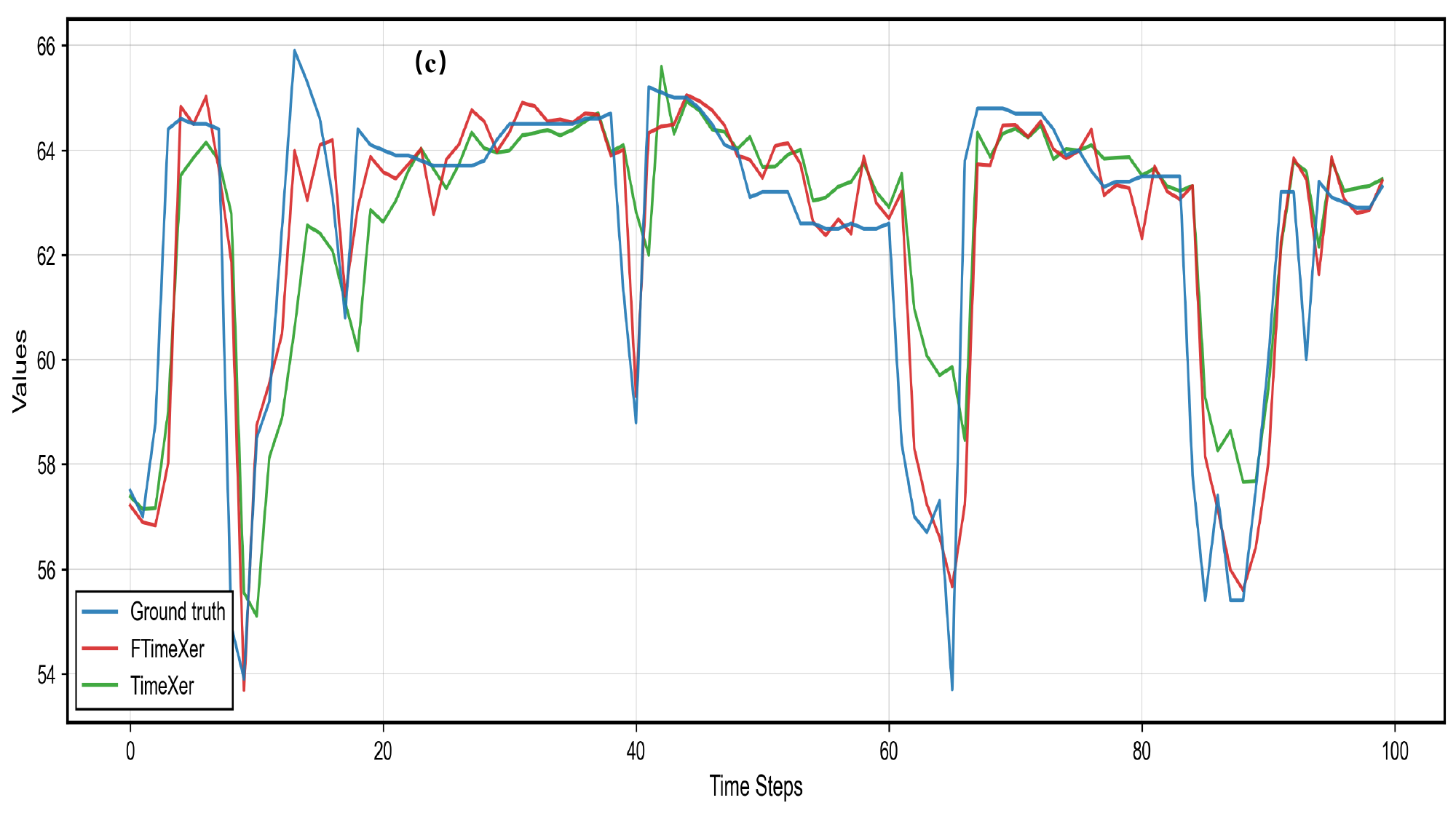}
  \caption{Visualization results on  Magnolia, California\_CT2, and NewYork\_Greenidge datasets.}
  \label{fig2:visualization}
\end{figure}

Figure~\ref{fig2:visualization} presents the visualization results across three datasets. Specifically, (a) corresponds to Magnolia, (b) corresponds to California\_CT2, and (c) corresponds to New York\_Greenidge. The blue line denotes the ground truth, the red line represents FTimeXer, and the green line indicates the performance of the second-best model, TimeXer. As illustrated in Fig.~\ref{fig2:visualization}, FTimeXer consistently captures peak and trough positions more accurately and tracks turning points more closely than TimeXer across all three datasets, indicating improved fidelity to rapid temporal fluctuations. This improvement stems from the FFT-based frequency branch and gated time--frequency fusion, which explicitly model periodic and oscillatory patterns while preserving sensitivity to rapid temporal variations. Moreover, the robust training strategy—random exogenous masking combined with consistency regularization—reduces spurious correlations arising from missing values and temporal misalignment in exogenous inputs, thereby yielding more stable predictions in real-world industrial deployments.

\section{CONCLUSION}
This paper investigates power-grid carbon footprint forecasting. We propose \textbf{FTimeXer}, a frequency-aware Transformer that incorporates an FFT-driven frequency branch and gated time--frequency fusion to more effectively capture multi-scale periodic and oscillatory patterns. In addition, to enhance robustness to imperfect exogenous signals, we introduce a backbone-agnostic training strategy that combines stochastic exogenous masking with consistency regularization. Extensive experiments on three real-world datasets demonstrate that FTimeXer delivers superior accuracy and robustness over representative baselines, thereby providing reliable predictive support for PCF accounting and emission-aware optimization. Building on these results, future work will investigate finer-grained uncertainty quantification and deployment-oriented adaptations for streaming scenarios and distribution-shift conditions.

\bibliographystyle{IEEEtran}
\bibliography{refs-lqz}

\end{document}